\documentclass[11pt]{article}
\usepackage{graphicx}
\usepackage{adjustbox} 
\usepackage{amsmath}
\usepackage{hyperref}
\usepackage{geometry}
\usepackage{fancyvrb}
\usepackage{xcolor}
\usepackage[most]{tcolorbox}
\usepackage{float}
\usepackage{multirow} 
\usepackage{fontawesome}
\usepackage[capitalize]{cleveref}
\usepackage{hhline}
\usepackage{makecell}
\usepackage{threeparttable}
\usepackage{appendix}

\definecolor{qual-fig-green}{RGB}{0,144,11}
\definecolor{qual-fig-red}{RGB}{238,0,0}
\definecolor{qual-fig-purple}{RGB}{153,51,255}

\definecolor{answergreen}{RGB}{222,255,222}
\definecolor{promptblue}{RGB}{235,245,255}   
\definecolor{promptgreen}{RGB}{235,255,235}  
\definecolor{promptgray}{RGB}{248,248,248}   

\newcommand{\bluearrow}[1]{(\textcolor{blue}{$\uparrow${#1}\%})}
\newcommand{\redarrow}[1]{(\textcolor{red}{$\downarrow${#1}\%})}

\newlength{\myindent}
\title{\textbf{Evaluating GPT-5 as a Multimodal Clinical Reasoner: A Landscape Commentary}}

\author{
Alexandru Florea$^{1, 2}$  \quad Shansong Wang$^{1}$  \quad Mingzhe Hu$^{1}$  \quad Qiang Li$^{1}$  \quad Zach Eidex$^{1}$ \\ \quad Luke del Balzo$^{1}$ \quad Mojtaba Safari$^{1}$ \quad {Xiaofeng Yang}$ \textsuperscript{1, 2 \faEnvelope}$ 
\vspace{3mm}
\\
$^1$Department of Radiation Oncology, Winship Cancer Institute,\\
Emory University School of Medicine\\
$^2$Department of Biomedical Engineering, Georgia Institute of Technology\\
\faEnvelope \quad Corresponding author: xiaofeng.yang@emory.edu \\
}
\date{\today}

\begin{document}
\maketitle

\begin{abstract}
The transition from task-specific artificial intelligence toward general-purpose foundation models raises fundamental questions about their capacity to support the integrated reasoning required in clinical medicine, where diagnosis demands synthesis of ambiguous patient narratives, laboratory data, and multimodal imaging. This landscape commentary provides the first controlled, cross-sectional evaluation of the GPT-5 family (GPT-5, GPT-5 Mini, GPT-5 Nano) against its predecessor GPT-4o across a diverse spectrum of clinically grounded tasks, including medical education examinations (USMLE), text-based reasoning benchmarks (MedQA, MedXpertQA Text), and visual question-answering in neuroradiology (BraTS cohorts), digital pathology (PathVQA, BreakHis), and mammography (EMBED, InBreast, CMMD, CBIS-DDSM) using a standardized zero-shot chain-of-thought protocol. GPT-5 demonstrated substantial gains in expert-level textual reasoning, with absolute improvements exceeding 25 percentage-points on MedXpertQA, establishing a robust foundation for clinical inference. When tasked with multimodal synthesis, GPT-5 effectively leveraged this enhanced reasoning capacity to ground uncertain clinical narratives in concrete imaging evidence, achieving state-of-the-art or competitive performance across most VQA benchmarks and outperforming GPT-4o by margins of 10-40\% in mammography tasks requiring fine-grained lesion characterization. However, performance remained moderate in neuroradiology (44\% macro-average accuracy) and lagged substantially behind domain-specific models in mammography, where specialized systems exceed 80\% accuracy compared to GPT-5's 52-64\%. These findings indicate that while GPT-5 represents a meaningful advance toward integrated multimodal clinical reasoning, mirroring the clinician's cognitive process of biasing uncertain information with objective findings, generalist models are not yet substitutes for purpose-built systems in highly specialized, perception-critical tasks. This commentary delineates the current landscape, positioning GPT-5 as a powerful adjunct capable of holistic reasoning while underscoring that domain adaptation, rigorous validation, and guarantees of reasoning transparency remain essential prerequisites for clinical deployment.
\end{abstract}

\section{Introduction}
Recent advancements in large language models (LLMs) have fostered a paradigm shift from narrowly-optimized, task-specific systems toward domain-invariant, general purpose models~\cite{achiam2023gpt,liu2024deepseek,openai2025gpt5}. In medicine, this transition has opened avenues permitting the integration of LLMs into clinical decision support systems, medical education, and healthcare administration~\cite{vrdoljakReviewLargeLanguage2025}. However, real-world clinical reasoning is not an isolated process; it requires the expert synthesis of a cohesive narrative from clinicopathologic components of evidence including patient histories, clinical exam findings, lab reports, microscopic findings, and imaging data. To help guide this judgment and propose appropriate treatment plans, domain knowledge must be organized into a structural framework that captures the relationships between seemingly independent parcels of data into a cohesive clinical picture~\cite{shinReasoningProcessesClinical2019}. For LLMs, this means not only understanding language but performing consistent reasoning and decision-making across heterogeneous modalities~\cite{liu2025application}.

Enabling LLMs to reliably perform this type of multimodal medical reasoning without relying on extensive domain-specific fine-tuning is becoming a key issue in medical artificial intelligence (AI)~\cite{nori2023capabilities,thirunavukarasu2023large,hu2024advancing}. Empirical studies across a range of clinical specialties including neurosurgery~\cite{hopkins2023chatgpt}, hepatology~\cite{yeo2023assessing}, and core internal medicine domains~\cite{johnson2023assessing} have exhibited that LLMs can achieve strong performance in knowledge recall and structured reasoning. Parallel efforts have explored their use as decision-support tools in image-centric domains such as radiology~\cite{bhayana2023gpt}, pathology~\cite{omar2024chatgpt}, and orthodontics~\cite{demir2024enhancing}. In routine clinical workflows, LLMs have further demonstrated utility in generating clinic reports~\cite{tung2024comparison}, discharge summaries~\cite{kweon2024ehrnoteqa}, and cancer screening plans~\cite{hao2025large}. Despite these promising results, most existing studies remain predominantly text-centric and employ heterogeneous methodologies. This obscures how these models' capabilities translate to settings that require joint reasoning over reports, images, and structured signals.

To examine this gap, we evaluated GPT-5~\cite{openai2025gpt5} as a generalist multimodal reasoner and investigated whether a single instruction-following model can reliably support multimodal clinical reasoning. We first evaluated GPT-5 on text-centric question-answering (QA) tasks to assess whether LLMs exhibit foundational competencies in reasoning over structured clinical information. Clinical decision-making frequently involves diagnostic uncertainty, particularly in settings where multiple conditions present with overlapping objective clinical findings, or in complex instances where definitive diagnostic tests may not currently exist, requiring diagnoses of exclusion. To evaluate model performance under these conditions, we subsequently assessed GPT-5 on multimodal visual question-answering (VQA) tasks that integrate textual and imaging evidence, which allowed for testing of the hypothesis that medical imaging can provide complementary information that helps resolve ambiguity in clinical narratives and supports more robust clinical conclusions. Finally, we perform comparisons with GPT-5’s developmental predecessor, GPT-4o~\cite{achiam2023gpt}, across all tasks to contextualize its advancements.

We focus our evaluation on three clinically important and methodologically distinct imaging domains: neuroradiology, digital pathology, and mammography. In neuroradiology, brain tumors, including glioblastoma multiforme (GBM), meningioma, and brain metastases, constitute a heterogeneous group of intracranial lesions with distinct radiographic features, biological behavior, and prognostic implications~\cite{mcfaline2018brain}. Accurate and timely differentiation of these pathologies is critical, as it directly informs prognostication and patient management~\cite{10.1093/neuonc/noab106}. Magnetic resonance imaging (MRI) is the current clinical gold standard for brain tumor evaluation due to its superior soft-tissue contrast and multiparametric capabilities; however, interpretation requires substantial subspecialty expertise, and diagnostic variability can persist even among experts~\cite{laustsen2025interobserver,crowe2017expertise}. This degree of single-modality uncertainty provides an ideal test case for evaluating whether an LLM can standardize interpretations by grounding a clinical history in objective imaging features.

Digital pathology represents a second, and fundamentally different, multimodal challenge for model evaluation. As a cornerstone of cancer diagnosis and management~\cite{sulaievaDigitalPathologyImplementation2024}, digital pathology relies on the analysis of high-resolution whole slide images (WSIs) that can span gigapixel scales, encompass multi-resolution contextual information, and exhibit significant stain variability and inter-institutional domain shift~\cite{campanella2019clinical,bejnordi2017diagnostic}. In hematopathology and cytology, diagnostic reasoning further depends on subtle morphological cues such as nuclear lobulation, chromatin texture, and cytoplasmic granularity~\cite{matek2019human}. These characteristics place substantial demands on models to integrate fine-grained visual perception with clinically grounded semantic reasoning in our text-based evaluations.

Finally, mammography comprises the third multimodal challenge. As the ubiquitous, population-level breast cancer screening tool, mammography represents a high-volume imaging modality, with a unique, seemingly near-binary clinical decision output: the recommendation to continue via follow-ups with imaging, or obtaining a sample of a suspicious lesion. Breast cancer is the most frequently diagnosed malignancy among women worldwide and a leading cause of cancer-related mortality~\cite{kim2025global}. Although mammography enables early detection of abnormal tissue changes, image interpretation is complex and resource-intensive, requiring specialized training and extensive clinical experience. Subtle early-stage findings, combined with wide variability in breast tissue composition, contribute to substantial interpretive difficulty~\cite{lehman2017national}. Even among expert radiologists, inter-reader variability remains significant~\cite{gilbert2006single}. These challenges make mammography a stringent test case for evaluating whether generalist multimodal LLMs can support clinically meaningful, cost effective, and potentially more consistent image-based reasoning.

Together, these domains provide a comprehensive and clinically grounded testbed for assessing the extent to which GPT-5 advances multimodal medical reasoning beyond prior-generation models.

\section{Methods}
\begin{figure}[h!]
    \centering
    \includegraphics[width=1\linewidth]{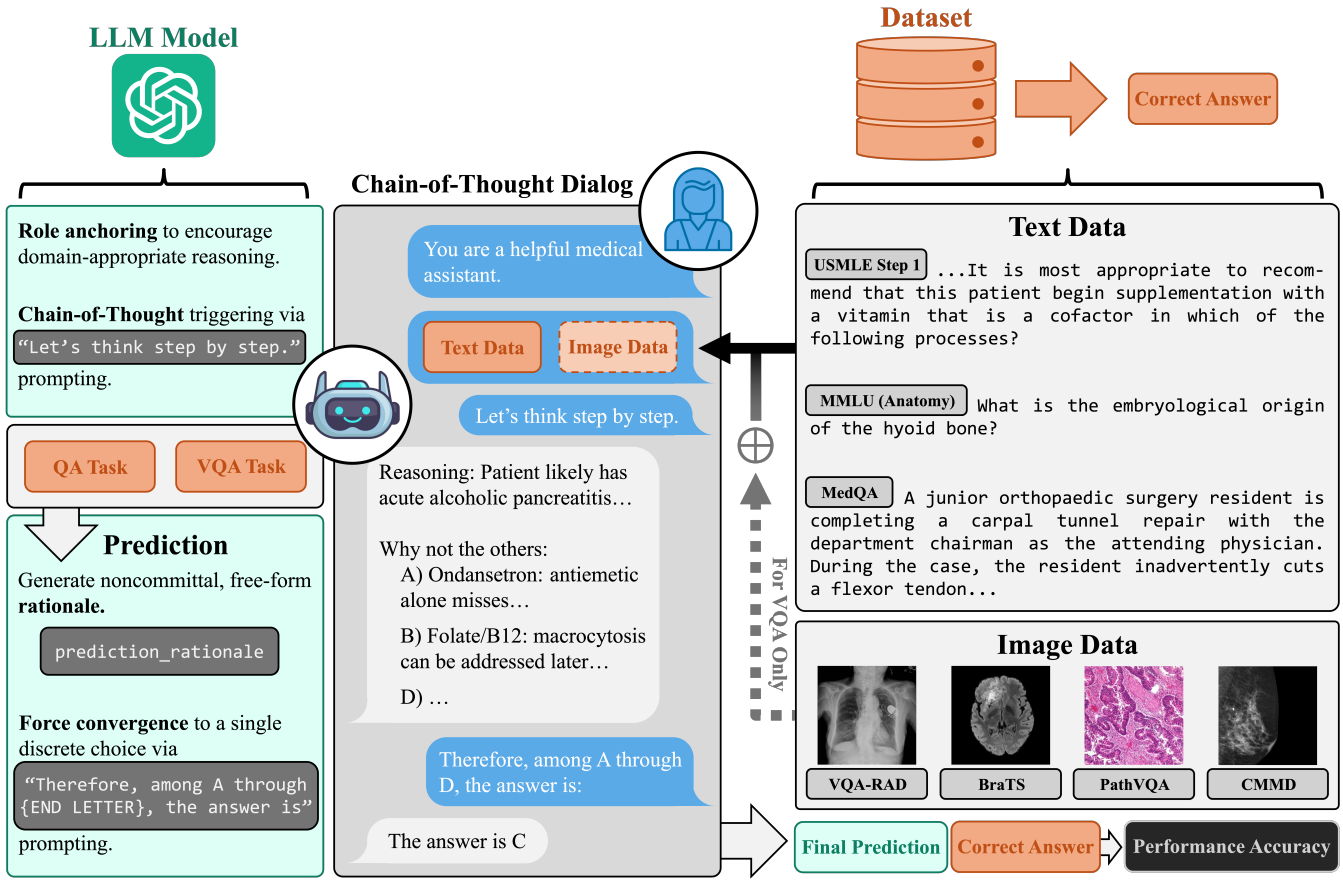}
    \vspace*{-6mm} 
    \caption{Workflow diagram of the proposed evaluation pipeline. (Right panel) Datasets are first standardized and relevant data is extracted into the chain-of-thought (CoT) prompting design. For Question Answering (QA) tasks, only textual data is passed to the CoT dialog; for Visual Question Answering (VQA) tasks, both text and image data is provided. The correct answer used for evaluating model performance is saved from the dataset. (Left panel) Each of four LLM models---GPT-5, GPT-5 Mini, GPT-5 Nano, and GPT-4o---are evaluated under a zero-shot protocol. First, the model is anchored to the role of a medical assistant to elucidate clinical thinking and reasoning. Then, a chain-of-thought reasoning scheme is triggered via prompting, and relevant question-answering data is passed from the dataset to the model for answering. This step enables a free-flowing thought process generation without the model deciding on a final answer choice. The model's prediction rational is saved and can be examined in detail. Finally, answer choice convergence is forced via direct prompting. This final choice is compared to the correct answer to assess performance accuracy (bottom right). Each model is exposed to the same inputs. (Middle panel) Chain-of-thought dialog example. User input is shown in blue, while LLM output is shown in grey.}
    \label{fig:workflow}
\end{figure}

This work evaluated GPT-5, GPT-5 Mini, GPT-5 Nano, and GPT-4o under a strict zero-shot protocol, with no fine-tuning and no in-context examples, on numerous benchmarks. We standardize splits and prompts across GPT models, evaluating zero-shot regimes with the same exemplars, chain-of-thought (CoT) supervision, and answers constrained to a single final choice for multiple-choice questions. This design explicitly isolates the contribution of the model-level advances from confounding factors such as prompt engineering or dataset-specific heuristics. An end-to-end workflow diagram is shown in Figure~\ref{fig:workflow}.

\subsection{Datasets}
We consider fourteen datasets that span both text-based and multimodal medical reasoning tasks. Together, these datasets cover a wide range of medical specialties, reasoning types, and input modalities.

To establish a foundational understanding of structured clinical information, we first evaluate performance on text-based, standardized medical education questions using the USMLE Self Assessment dataset~\cite{kung2023performance} drawn from official USMLE practice materials\footnote{\url{https://www.usmle.org/prepare-your-exam}} for Step 1, Step 2 Clinical Knowledge (CK), and Step 3. This setup follows the protocol of Nori et al.~\cite{nori2023capabilities}, and the original structure and content of the official sample materials are preserved. In addition, we used MedQA~\cite{jin2020disease}, which contains multiple-choice questions drawn from medical licensing examinations in the U.S., Mainland China, and Taiwan, provided in English, Simplified Chinese, and Traditional Chinese. The U.S. and Mainland China subsets use five options per question (1,273 and 3,426 questions in their respective test splits), while the Taiwan subset contains 1,413 questions with four options. Because a simplified U.S. split is available with four options per item (constructed by removing one incorrect choice), we evaluated on the simplified U.S. test set to align with a uniform multiple-choice interface. We additionally used the medical subset of MMLU~\cite{hendryckstest2021} to probe breadth across medical knowledge domains under a standardized multiple-choice format.

To assess expert-level clinical reasoning beyond broad-domain recall, we included MedXpertQA~\cite{zuo2025medxpertqa}, which comprises 4,460 questions spanning 17 specialties and 11 body systems and is partitioned into text-only (MedXpertQA Text) and multimodal (MedXpertQA MM) subsets; the dataset construction emphasizes clinical relevance and difficulty through board-style questions, rigorous filtering, synthesis intended to mitigate leakage, and multiple rounds of expert review.

For multimodal reasoning in radiology and related imaging settings, we evaluated VQA-RAD~\cite{lau2018dataset}, which includes 2,244 question--answer pairs linked to 314 radiology images sourced from the MedPix database and manually curated by clinicians. Because our primary evaluation uses discrete-choice scoring, we restricted VQA-RAD to the binary ``yes/no'' items in the test split (251 samples).

We further assessed neuro-oncology imaging reasoning using BraTS cohorts that provide multiparametric MRI for glioblastoma (GLI)~\cite{menze2014multimodal}, meningioma (MEN)~\cite{labella2023asnr}, and brain metastases (MET)~\cite{moawad2024brain}. For each case, four MRI sequences were available (T1c, T1w, T2w, and T2-FLAIR), along with tumor segmentation masks to localize lesions. Each imaging case was paired with radiology-style textual findings generated in the style of AutoRG-Brain or obtained from equivalent clinical reporting sources, enabling joint evaluation of image-grounded interpretation and clinically contextual reasoning.

For digital pathology and hematology-focused visual reasoning, we evaluated PathVQA~\cite{he2020pathvqa} using the closed-question subset of the original test split, with images stratified by source into whole slide images (WSIs), clinical/microscopic images, medical diagrams, and gross specimens. To support fair zero-shot evaluation, we standardized the answer space, normalized casing and synonyms, and removed ambiguous items. We also selected two complementary test subsets from OmniMedVQA~\cite{hu2024omnimedvqa}, a large-scale benchmark spanning multiple datasets and modalities: Blood Cell VQA (hematology), which assesses knowledge-grounded QA on peripheral blood cell images, and BreaKHis (breast histopathology), which probes zero-shot benign--malignant triage and morphological understanding on microscopy.

For mammography reasoning, we constructed and evaluated VQA-style tasks using four public datasets to cover variations in acquisition, labeling, and population characteristics. EMBED~\cite{jeong2023embed} contains 3.4 million screening and diagnostic images from 110,000 patients collected between 2013 and 2020 and emphasizes generalizability across ethnic groups through balanced representation of Black and White women. It includes 2D, synthetic 2D (C-view), and 3D digital breast tomosynthesis (DBT) images, with lesion annotations linked to structured descriptors and pathology-confirmed outcomes across six severity classes. In this study, we used the publicly available subset, corresponding to 20\% of the 2D portion. The InBreast~\cite{moreira2012inbreast} dataset comprises 115 breast cancer cases (410 images) spanning bilateral and mastectomy studies and includes heterogeneous lesion types such as masses, calcifications, asymmetries, and distortions. CMMD~\cite{cui2021cmmd} provides full-field digital mammography (FFDM) images from 1,775 patients collected between 2012 and 2016, accompanied by clinical metadata including age, lesion type, and pathology results, with breast-level benign/malignant labels based on pathology-confirmed outcomes. Finally, CBIS-DDSM~\cite{sawyer2016cbisddsm} contains 2,620 scanned film mammography studies with pathology-verified labels. From this dataset, we derived VQA items targeting lesion type classification, ROI-based localization and attributes, and pathology-related queries when available. To improve comparability across datasets and question types, we applied balanced sampling by selecting equal numbers of items per question and answer category, mitigating class imbalance while maintaining a tractable evaluation set size.

\subsection{Prompting Design} \label{sec:promptDesign}
All models were evaluated using a standardized CoT prompting protocol designed to elicit explicit intermediate reasoning while enforcing a discrete final answer format for reproducible scoring. Each sample was executed as a short, multi-turn chat. A system message anchors the medical assistant role to encourage domain-appropriate reasoning. The first user turn presents the question and explicitly triggers CoT via ``\textcolor{blue}{Let’s think step by step.}'' The assistant then produces a free-form rationale (stored as \textit{prediction\_rationale}) without committing to an option. 

To ensure consistent, machine-readable outputs across datasets and models, we introduced a second user turn that forced convergence to a single discrete choice: ``\textcolor{blue}{Therefore, among A through \textcolor{red}{\{END\_LETTER\}}, the answer is}'', where \textcolor{red}{\{END\_LETTER\}} denotes the last option letter computed from the number of choices. The final assistant turn returns the option letter (stored as \textit{prediction}). For multimodal items, all images associated with the sample are appended as \texttt{image\_url} entries to the first user message, enabling the model to reason over text and images within a single turn while keeping the subsequent convergence step purely textual. The JSON templates implementing the QA and VQA variants of this protocol are provided in Figures~\ref{fig:promptQA} and~\ref{fig:promptVQA}, and concrete instantiations are included in Appendix~\ref{sec:appendix}.

For neuroradiology VQA tasks, MRI volumes were converted to a standard Right, Anterior, Superior (RAS) orientation prior to analysis to ensure spatial consistency across subjects. Tumor centroids were calculated directly from the segmentation masks. From each volume, axial, sagittal, and coronal slices passing through the tumor centroid were extracted. Intensities were normalized using percentile-based clipping (2nd to 98th percentile) to suppress outliers and enhance lesion visibility. The three slices were concatenated to form a single triplanar mosaic image, providing a compact but information-rich visual representation for subsequent evaluation. The accompanying clinical text for each case was processed using a custom parsing pipeline to extract structured descriptors of tumor appearance. This process identified the primary lesion location, enhancement pattern, boundary definition, the presence of peritumoral edema, the presence or absence of midline shift, maximum lesion dimension in millimeters, involvement across the midline, and compression of any ventricular structures. The parser combined rule-based term matching, synonym normalization, and regular expressions for numerical values to ensure consistent extraction across reports. These structured features were then transformed into VQA items following the style protocol described in Section~\ref{sec:promptDesign}. Each feature was mapped to a natural-language question designed to assess either visual understanding or clinical reasoning. For example, enhancement features yielded yes/no questions such as “Does the lesion show ring enhancement?”, location features were converted into multiple-choice questions with plausible distractors, and lesion size measurements were framed as numeric multiple-choice questions. Each VQA item included the question text with labeled answer choices, the correct answer, the associated triplanar mosaic, and metadata specifying the medical task, body system, and question type.

For digital pathology and hematology VQA tasks (PathVQA, Blood Cell VQA, and BreaKHis), we used the same multimodal prompting template shown in Figure~\ref{fig:promptVQA} directly on the benchmark images and associated questions, after dataset-specific answer-space standardization and filtering.

For each mammography dataset, VQA questions were automatically generated from the structured panel data and metadata accompanying the associated image, including patient information, imaging annotations, lesion type, BI-RADS density category, and biopsy-confirmed pathology. This design yields questions that target well-defined clinical tasks, such as ``what is the BI-RADS breast density?'' or ``is the imaging finding suggestive of malignancy?'', ensuring a one-to-one correspondence between the query, its answer, and the correct label. By avoiding free-text questions based on subjective interpretation of image texture, morphology, or density patterns, the approach reduces variability and enhances reproducibility in automated evaluation. However, the absence of texture-oriented, descriptive queries limits the direct assessment of visual reasoning processes, positioning the evaluation as a test of structured clinical label inference rather than nuanced image-based reasoning.

\begin{figure}[H]
    \centering
    \begin{tcolorbox}[
            colback=promptgray,
            colframe=black!50!black,
            colbacktitle=white!50!black,
            title=\textbf{Question-Answering Prompt Design},
            fonttitle=\Large,
            fontupper=\large,
            center title,
            toptitle=8pt,
            bottomtitle=5pt,
            left=-20pt,
            bottom=-5pt,
        ]
        
    {\linespread{0.85}\selectfont
    \begin{verbatim}
    [
      {
        "role": "system",
        "content": "You are a helpful medical assistant."
      },
      {
        "role": "user",
        "content": [
          {
            "type": "text",
            "text": "Q: {QUESTION_TEXT}\nA: Let's think step by step."
          }
        ]
      },
      {
        "role": "assistant",
        "content": "{ASSISTANT_RATIONALE}"
      },
      {
        "role": "user",
        "content": "Therefore, among A through {END_LETTER}, the answer is"
      },
      {
        "role": "assistant",
        "content": "{ASSISTANT_FINAL}"
      }
    ]
    \end{verbatim}
    }
    \end{tcolorbox}

    \vspace*{-3mm} 
    \caption{Zero-shot chain-of-thought prompting template for text-only question-answering (QA) tasks, consisting of an initial rationale-generation turn followed by a constrained, single-letter answer selection.}
    \label{fig:promptQA}
\end{figure}

\newcommand*{\fvtextcolor}[2]{\textcolor{#1}{#2}}

\begin{figure}[H]
    \centering
    \begin{tcolorbox}[
            colback=promptgray,
            colframe=black!50!black,
            colbacktitle=white!50!black,
            title=\textbf{Visual Question-Answering Prompt Design},
            fonttitle=\Large,
            fontupper=\large,
            center title,
            toptitle=8pt,
            bottomtitle=5pt,
            left=-20pt,
        ]
        
    {\linespread{0.85}\selectfont
    \begin{Verbatim}[commandchars=+\(\)]
    [
      {
        "role": "system",
        "content": "You are a helpful medical assistant."
      },
      {
        "role": "user",
        "content": [
          {
            "type": "text",
            "text": "Q: {QUESTION_TEXT}\nA: Let's think step by step."
          },
          +fvtextcolor(red)({)
            +fvtextcolor(red)("type": "image_url",)
            +fvtextcolor(red)("image_url": {"url": "{IMAGE_URL_1}"})
          +fvtextcolor(red)(})
        ]
      },
      {
        "role": "assistant",
        "content": "{ASSISTANT_RATIONALE}"
      },
      {
        "role": "user",
        "content": "Therefore, among A through {END_LETTER}, the answer is"
      },
      {
        "role": "assistant",
        "content": "{ASSISTANT_FINAL}"
      }
    ]
    \end{Verbatim}
    }
    \end{tcolorbox}

    \vspace*{-3mm} 
    \caption{Zero-shot chain-of-thought prompting template for multimodal visual question-answering (VQA) tasks. An initial rationale-generation turn is accompanied with a piece of imaging evidence and followed by a constrained, single-letter answer selection.}
    \label{fig:promptVQA}
\end{figure}

\section{Results}
All results are reported in Table~\ref{tbl:results} as the percentage of correct answers, unless otherwise stated. All comparisons are made using absolute percentage-point differences.

\begin{table}[htbp]
    \centering
    \small
    \vspace*{-0.45in}
    \caption{Model performance (\%) on QA and VQA datasets. The blue or red numbers and arrows indicate absolute percentage-point increases or decreases, respectively, of GPT-5 when compared to GPT-4o. Boldface results indicate the highest performing model.}
    \label{tbl:results}
        \adjustbox{center}{
            \begin{tabular}{lcccc}
            \hhline{=====}
            \textbf{Dataset} & \textbf{GPT-5} & \textbf{GPT-5 Mini} & \textbf{GPT-5 Nano} & \textbf{GPT-4o} \\
            \hline
            \multicolumn{5}{l}{\textbf{USMLE}}\\
            \hspace{\myindent} Step 1 & \textbf{93.28} \bluearrow{0.84} & 93.28 & 93.28 & 92.44 \\
            \hspace{\myindent} Step 2 & \textbf{97.50} \bluearrow{4.17} & 95.83 & 90.00 & 93.33 \\
            \hspace{\myindent} Step 3 & \textbf{94.89} \bluearrow{3.65} & 94.89 & 92.70 & 91.24 \\
            \emph{Macro-Average}      & \textbf{95.22} \bluearrow{2.88} & 94.67 & 91.99 & 92.34 \\
            \hline
            \multicolumn{5}{l}{\textbf{MedQA}}\\
            \hspace{\myindent} US (4-Option) & \textbf{95.84} \bluearrow{4.80} & 93.48 & 91.44 & 91.04 \\
            \hline
            \multicolumn{5}{l}{\textbf{MMLU}}\\
            \hspace{\myindent} Anatomy & \textbf{92.59} \bluearrow{1.48} & 92.59 & 88.15 & 91.11 \\
            \hspace{\myindent} Clinical Knowledge & \textbf{95.09} \bluearrow{2.64} & 91.32 & 89.81 & 92.45 \\
            \hspace{\myindent} College Biology & \textbf{99.31} \bluearrow{2.09} & 99.31 & 97.92 & 97.22 \\
            \hspace{\myindent} College Medicine & \textbf{91.91} \bluearrow{1.74} & 88.44 & 85.55 & 90.17 \\
            \hspace{\myindent} Medical Genetics & \textbf{100.00} \bluearrow{4.00} & 99.00 & 98.00 & 96.00 \\
            \hspace{\myindent} Professional Medicine & \textbf{97.79} \bluearrow{1.10} & 97.43 & 96.69 & 96.69 \\
            \hline
            \multicolumn{5}{l}{\textbf{MedXpertQA Text}}\\
            \hspace{\myindent} Reasoning & \textbf{56.96} \bluearrow{26.33} & 45.94 & 36.38 & 30.63 \\
            \hspace{\myindent} Understanding & \textbf{54.84} \bluearrow{25.30} & 43.80 & 33.96 & 29.54 \\
            \hline
            \multicolumn{5}{l}{\textbf{MedXpertQA MM}}\\
            \hspace{\myindent} Reasoning     & \textbf{69.99} \bluearrow{29.26} & 60.51 & 45.44 & 40.73 \\
            \hspace{\myindent} Understanding & \textbf{74.37} \bluearrow{26.18} & 61.37 & 45.85 & 48.19 \\
            \hline
            \multicolumn{5}{l}{\textbf{VQA-RAD}}\\
            \hspace{\myindent} Specific Question & 70.92 \bluearrow{1.01} & \textbf{74.90} & 65.34 & 69.91 \\
            \hline
            \multicolumn{5}{l}{\textbf{BraTS}}\\
			\hspace{\myindent} MET     & \textbf{42.68} \bluearrow{4.20} & 42.09 & 35.95 & 38.48 \\
			\hspace{\myindent} GLI     & 46.34 \redarrow{3.46} & 48.97 & 38.00 & \textbf{49.80} \\
			\hspace{\myindent} MEN     & \textbf{42.12} \bluearrow{5.93} & 41.52 & 33.60 & 36.19 \\
			\emph{Macro-Average} & 43.71 \bluearrow{2.22} & \textbf{44.19} & 35.85 & 41.49 \\
            \hline
            \multicolumn{5}{l}{\textbf{PathVQA}}\\
            \hspace{\myindent} WSI      & \textbf{69.4} \bluearrow{1.37} & 68.6 & 62.3 & 68.03 \\
            \hspace{\myindent} Diagram  & \textbf{76.7} \bluearrow{2.5} & 72.5 & 71.1 & 74.2 \\
            \hspace{\myindent} Gross    & \textbf{70.3} \bluearrow{1.6} & 69.7 & 65.8 & 68.7 \\
            \hspace{\myindent} Clinical & \textbf{78.9} \bluearrow{5.75} & 74.7 & 68.1 & 73.15 \\
            \emph{Weighted Accuracy}    & \textbf{70.9} \bluearrow{1.9} & 69.6 & 63.9 & 69.0 \\
            \hline
            \multicolumn{5}{l}{\textbf{Blood Cell VQA}}\\
            \hspace{\myindent} Specific Question   & 75.2 \redarrow{4.8}& 74.2 & 55.3 & \textbf{80.0} \\
            \hline
            \multicolumn{5}{l}{\textbf{BreaKHis}}\\
            \hspace{\myindent} Specific Question & \textbf{71.7} \bluearrow{0.2} & 61.2 & 46.0 & 71.5 \\
            \hline
            \multicolumn{5}{l}{\textbf{EMBED}}\\
            \hspace{\myindent} Density       & \textbf{56.8} \bluearrow{32.5} & 34.3 & 24.8 & 24.3 \\
            \hspace{\myindent} Distortion    & 52.5 \bluearrow{32.5} & \textbf{53.5} & \textbf{53.5} & 20.0 \\
            \hspace{\myindent} Mass          & \textbf{64.5} \bluearrow{14.5} & 60.8 & 52.3 & 50.0 \\
            \hspace{\myindent} Calcification & \textbf{63.5} \bluearrow{19.2} & 57.3 & 51.5 & 44.3 \\
            \hspace{\myindent} Malignancy    & \textbf{52.8} \bluearrow{10.3} & 47.3 & 47.8 & 42.5 \\
            \hline
            \multicolumn{5}{l}{\textbf{InBreast}}\\
            \hspace{\myindent} BI-RADS       & \textbf{36.9} \bluearrow{13.2} & 28.1 & 17.6 & 23.7 \\
            \hspace{\myindent} Abnormality   & 45.9 \bluearrow{9.8} & \textbf{49.2} & 37.8 & 36.1 \\
            \hspace{\myindent} Malignancy    & 35.0 \bluearrow{4.6} & \textbf{40.0} & 21.5 & 30.4 \\
            \hline
            \multicolumn{5}{l}{\textbf{CMMD}}\\
            \hspace{\myindent} Abnormality   & 32.3 \bluearrow{6.2} & \textbf{42.3} & 34.0 & 38.5 \\
            \hspace{\myindent} Malignancy    & 55.0 \bluearrow{6.5} & \textbf{63.3} & 52.7 & 48.5 \\
            \hline
            \multicolumn{5}{l}{\textbf{CBIS-DDSM}}\\
            \hspace{\myindent} BI-RADS       & \textbf{69.3} \bluearrow{40.9} & 43.6 & 20.2 & 28.4 \\
            \hspace{\myindent} Abnormality   & \textbf{66.0} \bluearrow{19.8} & 53.3 & 41.8 & 46.2 \\
            \hspace{\myindent} Malignancy    & \textbf{58.2} \bluearrow{18.2} & 43.5 & 39.0 & 40.0 \\
            \hhline{=====}
        \end{tabular}}
\end{table}

\subsection{Performance of GPT-5 on Medical Education Exams}
We begin this section by examining the GPT family of models on medical examination questions. Table \ref{tbl:results} reports accuracy on all three Step exams of the USMLE. GPT-5 outperformed all other models on all three USMLE board exams, with the largest margin on Step 2 CK (+4.17\%). Step 2 CK focuses on clinical decision-making and management, which is consistent with GPT-5’s improved CoT reasoning. The macro-average, an unweighted mean over all three Step exams, reached 95.22\% (+2.88\% vs GPT-4o), demonstrating the model’s readiness for high-stakes clinical reasoning tasks.

\subsection{Performance of GPT-5 on QA Benchmarks}
On text-based medical QA datasets, GPT-5 achieved consistent gains over GPT-4o and smaller GPT-5 variants. On MedQA (US 4-Option), GPT-5 reached 95.84\% accuracy, a 4.80 percentage-point improvement over GPT-4o, indicating stronger factual recall and diagnostic reasoning in clinical question contexts. The most pronounced gains appeared in MedXpertQA Text, where reasoning accuracy improved by 26.33\% and understanding by 25.30\% over GPT-4o. This suggests a substantial enhancement in multi-step inference and nuanced comprehension of medical narratives.
In MMLU medical subdomains, GPT-5 maintained near-ceiling performance ($>$91\% across all subjects), with notable gains in Medical Genetics (+4.00\%) and Clinical Knowledge (+2.64\%). The improvements were generally incremental in high-baseline categories, indicating that GPT-5 demonstrates larger gains on tasks with higher reasoning complexity, rather than purely factual recall.

\subsection{Performance of GPT-5 on VQA Benchmarks}
\subsubsection{Expert-Level Medical Reasoning and Understanding}
For multimodal reasoning, GPT-5 achieved a dramatic leap in MedXpertQA MM, with reasoning and understanding gains of +29.26\% and +26.18\%, respectively, relative to GPT-4o. This magnitude of improvement suggests significantly enhanced integration of visual and textual cues. However, in VQA-RAD, GPT-5 scored 70.92\%, slightly below GPT-5 Mini (74.90\%). Given VQA-RAD’s relatively small scale and radiology-specific nature, this difference may reflect benchmark-specific variability in a small test set or differences in answer calibration across models. 

A representative example from the MedXpertQA MM benchmark (Figure~\ref{fig:medPromptVQAsampleAnswer}) illustrates GPT-5’s capability to synthesize multimodal information in a clinically coherent manner. In this case, the model correctly identified esophageal perforation (Boerhaave syndrome) as the most likely diagnosis based on the combination of CT imaging findings, laboratory values, and key physical signs (suprasternal crepitus, blood-streaked emesis) following repeated vomiting. 
It then recommended a Gastrografin swallow study as the next management step, while explicitly ruling out other options and justifying each exclusion. This output demonstrates the model’s ability to integrate visual evidence with complex narrative context, maintain a structured diagnostic reasoning chain, and arrive at a high-stakes clinical decision that aligns with expert consensus.

\subsubsection{Brain Tumor MRI}
Across the three BraTS cohorts (MET, GLI, MEN), model accuracies clustered in a narrow band, with absolute values ranging from roughly one-third to one-half. The macro-average showed that GPT-5 achieved 43.71\% accuracy, GPT-5 Mini 44.19\%, GPT-5 Nano 35.85\%, and GPT-4o 41.49\%. Thus, the highest macro-average was observed for GPT-5 Mini, but its advantage over GPT-5 was 0.48 percentage points, which is small in absolute terms. Interestingly, GPT-4o achieved the best GLI accuracy (49.80\%), whereas GPT-5 led on MET (42.68\%) and MEN (42.12\%). GPT-5 Nano underperformed consistently in all cohorts.

\subsubsection{Digital Pathology}
On BreaKHis, GPT-5 attains 71.7\%, essentially tied with GPT-4o (71.5\%) and above GPT-5 Mini (61.2\%) and GPT-5 Nano (46.0\%). On PathVQA, GPT-5 yields the best weighted accuracy 70.9\% (vs. Mini 69.6\%, GPT-4o 69.0\%, Nano 63.9\%) and leads or matches across sources (WSI 69.4\%, diagram 76.7\%, gross 70.3\%, clinical 78.9\%), suggesting better cross-source consistency. In contrast, on Blood Cell VQA, GPT-4o is strongest (80.0\%), with GPT-5 close (75.2\%), GPT-5 Mini is comparable (74.2\%), and GPT-5 Nano is markedly lower (55.3\%). Taken together, these patterns indicate that fine-grained histopathology judgments benefit from larger model capacity. Furthermore, GPT-5’s advantage on PathVQA likely reflects improved robustness to source/format variability, however, hematology cell-type reasoning still favors GPT-4o, possibly due to cytology-specific cues (nuclear lobulation, granularity) and lexicon coverage.

\subsubsection{Mammography}
On the EMBED dataset, GPT-5 achieves the highest performance among GPT variants across all tasks (Density: 56.8\%, Distortion: 52.5\%, Mass: 64.5\%, Calcification: 63.5\%, Malignancy: 52.8\%), outperforming GPT-5 Mini, GPT-5 Nano, and GPT-4o. On InBreast, GPT-5 attains 36.9\% BI-RADS accuracy, 45.9\% for abnormality detection, and 35.0\% for malignancy classification. Interestingly, on CMMD, GPT-5 Mini achieves 42.3\% accuracy on abnormality detection and 63.3\% on malignancy classification, outperforming GPT-5 by 10.0\% and 8.3\%, respectively. On CBIS-DDSM, GPT-5 achieves 69.3\% BI-RADS accuracy, 66.0\% abnormality detection, and 58.2\% malignancy accuracy, exceeding GPT-4o performance. Across datasets, GPT-5's advantages are most pronounced in tasks requiring fine-grained lesion characterization, such as mass and calcification detection, and in multi-class tasks like BI-RADS classification.

To contextualize these results within the broader landscape, we compare them to state-of-the-art, domain-specific models. On EMBED, GPT-5's 52.8\% malignancy detection accuracy lags behind Mammo-CLIP's reported 82.3\%~\cite{chen2024mammo}. Similarly, on InBreast, GPT-5's 36.9\% BI-RADS accuracy is substantially lower than the 90.6\% achieved by specialized multi-scale networks~\cite{sun2025multi}.

\section{Discussion}
This study provides an evaluation of the GPT-5 family of models on a wide range of multimodal, domain-specific tasks particularly relevant to foundational medical education, clinical radiology, radiation oncology, and medical physics. Across text-based question answering (QA), visual question answering (VQA), and medical examination–style assessment, GPT-5 consistently demonstrated higher accuracy than its predecessors and compact variants, indicating substantial improvements in multimodal reasoning, clinical concept grounding, and numerical problem-solving. Nevertheless, several findings warrant careful interpretation, and the results reveal both strengths and limitations.

GPT-5 delivers substantial gains in multimodal medical reasoning, especially on datasets like MedXpertQA MM, EMBED, and CBIS-DDSM that demand tight integration of image-derived evidence with textual patient data. On these datasets, GPT-5 saw an absolute improvement of roughly 10\% to 40\% over GPT-4o in multimodal settings, consistent with improved cross-modal integration. This may reflect architectural refinements and expanded training corpora, which have meaningfully improved domain reasoning capacity. We observe that accuracy gains are most pronounced in reasoning-intensive tasks, in which CoT prompting likely synergizes with GPT-5’s enhanced internal reasoning capacity, enabling more accurate multi-hop inference. In comparison to domains where GPT-4o accuracy is already high (e.g. MMLU, BraTS, PathVQA), we note smaller or mixed differences, indicating that GPT-5's primary strength lies in its ability to tackle complex reasoning challenges rather than simply recalling facts.

On text-based tasks such as QA benchmarks and medical education questions, all four models demonstrated strong performance. Overall, GPT-5 earned consistent accuracy gains from 1\% to 7\% across most datasets, with a notably larger improvement of up to 26\% on MedXpertQA Text. This substantial improvement merits particular attention, and is hypothesized to reflect enhanced textual reasoning capabilities in GPT-5 relative to GPT-4o, which excelled at recall oriented question-answering but struggled to reason through more complicated, multi-layered questions. Conversely, GPT-5 demonstrated a greater capacity to reason through expert-level medical questions.

An unexpected observation was that GPT-5 scored slightly lower on VQA-RAD, BraTS, EMBED, InBreast, and CMMD compared to its smaller counterpart, GPT-5 Mini. We report differences of approximately 0.5\% to 10\% on the aforementioned datasets. The marginal differences between GPT-5 and GPT-5 Mini suggest that factors beyond model scale may influence performance in these domains. This discrepancy may be attributed to scaling-related differences in reasoning calibration; larger models might adopt a more cautious approach in selecting answers for smaller datasets, resulting in fewer, albeit more conservative, correct predictions. Future research could explore adaptive prompting or calibration techniques specifically tailored for small-domain multimodal tasks.

While providing a clear measure of generalization, this study’s zero-shot evaluation protocol does not account for domain-specific adaptation. It remains possible that targeted fine-tuning, prompt engineering, or retrieval-augmented generation could close the gap for smaller models or further extend GPT-5’s lead. For example, our evaluation indicated that GPT-5 (while outperforming GPT-4o) struggled to achieve high accuracies on complex, domain-specific queries in mammography VQA. This limitation is likely due to the model’s lack of fine-tuning on high-resolution grayscale breast imaging and the absence of explicit adaptation to structured clinical tasks. For comparison, current state-of-the-art models perform significantly better. On the EMBED dataset, Mammo-CLIP~\cite{chen2024mammo} was able to achieve 82.3\% accuracy on malignancy classification. Similarly, the multi-scale region selection network of Sun et al.~\cite{sun2025multi} achieved 90.6\% on the InBreast dataset, HybMNet~\cite{chen2025enhancing} achieved 79.7\% on the CMMD dataset, and PHYSnet~\cite{lopez2022multi} achieved 82.0\% on the CBIS-DDSM dataset. These comparisons should be interpreted cautiously because the specialized models were trained and evaluated under task-specific pipelines that are not identical to our zero-shot VQA formulation. As established, consistent mammogram interpretation itself is inherently challenging, even for experienced breastradiologists. Diagnostic decisions often hinge on subtle, low-contrast features, such as microcalcification clusters, spiculated mass margins, or architectural distortions that can be obscured by heterogenous overlapping fibroglandular tissue, and further complicated by variations in acquisition parameters and image quality. Accurate assessment requires not only fine-grained visual discrimination, but also the integration of patient history, comparison with prior imaging records, and adherence to standardized reporting frameworks. These factors underscore why mammographic VQA is especially difficult for general-purpose AI systems without targeted domain adaptation and large-scale, modality-specific training. While generalist models are rapidly improving, they are not yet a substitute for purpose-built AI in highly specialized, perception-critical tasks like mammography. Their current value lies in extending their broad, integrated reasoning, not in matching the precision of dedicated systems.

Several avenues emerge for extending this work. First, a critical barrier to deployment remains the absence of guarantees around factual correctness and reasoning transparency. A large language model used in clinical practice should be able to reason through difficult questions, not just rely on exploiting patterns found in its training data. One way to assess this is by adopting the approach of Bedi et al.~\cite{bediFidelityMedicalReasoning2025}. In their work, the MedQA dataset was modified so the correct answer choice was replaced with ``None of the other answers'' (NOTA). If models truly reasoned through medical questions, accuracy should not vary significantly because of this change. However, the researchers showed that GPT-4o performance degraded from 85.29\% accuracy on the original dataset, to 48.53\% on the NOTA-modified dataset, an absolute drop of 36.76\%~\cite{bediFidelityMedicalReasoning2025}. Such experiments illustrate the fragility of LLMs, and highlight a need for explainable clinical decision support systems whose uncertainty can be quantified if they are ever to be reliably deployed into clinical workflows. Ultimately, clinical conclusions from LLMs in their current state can best serve physicians when applied to scenarios with a complementary pre-test probability to serve as contextual framework for result interpretation. 

Second, data leakage remains a concern for any model trained on a corpus of publicly available data. This is an issue for both zero-shot and fine-tuned models, as it artificially inflates performance on problems it might otherwise have seen during training. The apparent reasoning task can regress to a recollection of something it has seen during training. We hypothesize fine-tuning the GPT-5 family of models may improve performance on non-public datasets. Finally, accuracy was the sole evaluation metric used in this work. It will be instructive to examine additional important performance dimensions, such as reasoning transparency, uncertainty calibration, and error analysis.

\section{Conclusion}
This study presents the first controlled, comparative evaluation of GPT-5’s capabilities in multimodal medical reasoning, comparing its performance to GPT-4o and smaller GPT-5 variants under standardized zero-shot CoT prompting. Across diverse QA and VQA benchmarks, GPT-5 demonstrated substantial diagnostic and etiological knowledge. GPT-5's capacity to refine conclusions based on subjective evidence via the introduction of imaging evidence in expert-level tasks is markedly stronger than the older GPT-4o model. We have shown that the GPT-5 family of models is capable of integrating complex textual and visual information streams to produce accurate, clinically coherent responses to expert-level medical problems. In clinical applications, GPT-5 achieved moderate accuracy on a clinically grounded brain-tumor VQA benchmark; moderate, but task-sensitive, performance on digital pathology tasks; and considerable improvement on mammography tasks. Although not quite ready for independent clinical use, GPT-5 has shown appreciable improvement over earlier models, establishing LLMs as powerful tools to potentially augment patient care in clinical tasks when paired with domain adaptation, rigorous validation, and uncertainty-aware evaluation.

\section{Acknowledgments} This research is supported in part by the National Institutes of Health under Award Number R01CA272991, R01EB032680 and U54CA274513.

\section{Code Availability} The evaluation pipeline code, including prompting design and GPT-5 queries, source code used for inference, and all datasets used, has been made publicly available and can be accessed at \url{https://github.com/alexcflorea/GPT-5-Evaluation}.

\newpage
\appendix
\renewcommand\thefigure{\thesection\arabic{figure}} 
\setcounter{figure}{0}
\section{Appendix - Detailed VQA Prompts} \label{sec:appendix}

\begin{figure}[H]
    \centering
    \begin{tcolorbox}[
            colback=promptgray,
            colframe=black!50!black,
            colbacktitle=white!50!black,
            title=\textbf{Prompting Sample from BraTS-GLI (Case: BraTS-GLI-00017-000-t1n)},
            enhanced,
            overlay={
                \node[anchor=north east, inner sep=0pt, yshift=-31pt] 
                at (frame.north east) {\includegraphics[width=0.5\textwidth]{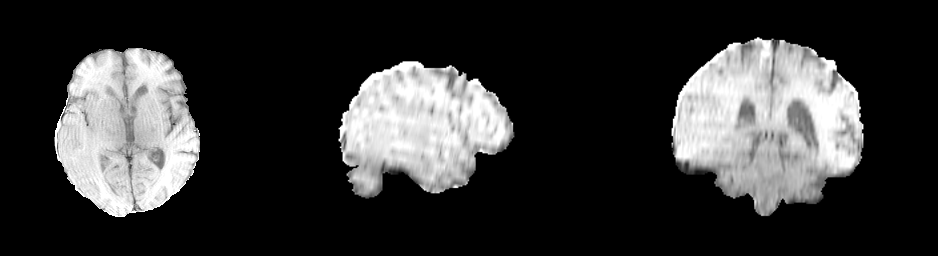}};
            } 
            fonttitle=\Large,
            fontupper=\large,
            center title,
            toptitle=8pt,
            bottomtitle=5pt,
            left=-20pt,
            bottom=-5pt,
            top=40pt
        ]
        
    {\linespread{0.85}\selectfont
    \begin{verbatim}
    [
      {
        "role": "system",
        "content": "You are a helpful medical assistant."
      },
      {
        "role": "user",
        "content": [
          {
            "type": "text",
            "text": "
                Q: Are the lesion boundaries clear or ill-defined         
                Answer Choices: 
                    (A) clear               (B) ill-defined 
                    (C) mixed/ambiguous       
                A: Let's think step by step."
          },
          {
            "type": "image_url",
            "image_url": {"url": "slices_out_gl/BraTS-GLI-00017-000-t1n_tri.png"}
          }
        ]
      },
      {
        "role": "assistant",
        "content": "{ASSISTANT_RATIONALE}"
      },
      {
        "role": "user",
        "content": "Therefore, among A through C, the answer is"
      },
      {
        "role": "assistant",
        "content": "B"
      }
    ]
    \end{verbatim}
    }
    \end{tcolorbox}

    \vspace*{-3mm} 
    \caption{Example VQA prompt for a representative BraTS-GLI case (Case: BraTS-GLI-00017-000-t1n). The triplanar mosaic image is provided first, followed by the question and labeled answer choices. The model outputs an intermediate reasoning step (\texttt{ASSISTANT\_RATIONALE}) and then a final letter (\texttt{ASSISTANT\_FINAL}) corresponding to its chosen answer. Accuracy was computed solely from \texttt{ASSISTANT\_FINAL}.}
    \label{fig:brainPromptVQAsample}
\end{figure}

\begin{figure}[H]
    \centering
    \begin{tcolorbox}[
            colback=promptgray,
            colframe=black!50!black,
            colbacktitle=white!50!black,
            title=\textbf{Prompting Sample from EMBED (Case: MM-4307)},
            enhanced,
            overlay={
                \node[anchor=north east, inner sep=0pt, yshift=-31pt] 
                at (frame.north east) {\includegraphics[width=0.25\textwidth]{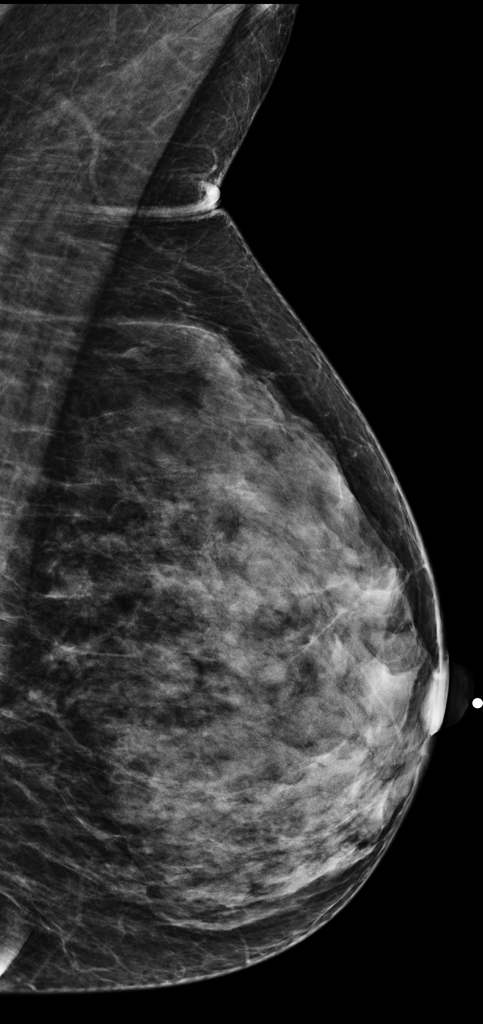}};
            } 
            fonttitle=\Large,
            fontupper=\large,
            center title,
            toptitle=8pt,
            bottomtitle=5pt,
            left=-20pt,
            bottom=-5pt,
        ]
        
    {\linespread{0.85}\selectfont
    \begin{verbatim}
    [
      {
        "role": "system",
        "content": "You are a helpful medical assistant."
      },
      {
        "role": "user",
        "content": [
          {
            "type": "text",
            "text": "
                Q: What is the BI-RADS breast density of 
                this image?          
                Answer Choices: 
                    (A)BI-RADS A            (B)BI-RADS B 
                    (C)BI-RADS C            (D)BI-RADS D
                A: Let's think step by step."
          },
          {
            "type": "image_url",
            "image_url": { "url": "images/MM-4307.png" }
          }
        ]
      },
      {
        "role": "assistant",
        "content": "{ASSISTANT_RATIONALE}"
      },
      {
        "role": "user",
        "content": "Therefore, among A through D, the answer is"
      },
      {
        "role": "assistant",
        "content": "D"
      }
    ]
    \end{verbatim}
    }
    \end{tcolorbox}

    \vspace*{-3mm} 
    \caption{Example VQA prompt for a representative EMBED case (Case: MM-4307). The associated image is provided first, followed by the question and labeled answer choices. The model outputs an intermediate reasoning step (\texttt{ASSISTANT\_RATIONALE}) and then a final letter (\texttt{ASSISTANT\_FINAL}) corresponding to its chosen answer. Accuracy was computed solely from \texttt{ASSISTANT\_FINAL}.}
    \label{fig:mamPromptVQAsample}
\end{figure}

 Representative samples and prompting designs from selected datasets are illustrated in Figures~\ref{fig:medPromptVQAsample} through \ref{fig:mamPromptVQAsample}. In Figure~\ref{fig:medPromptVQAsample} a detailed question from expert level VQA tasks is shown along with its associated imaging evidence. The corresponding reasoning and answer for that question is presented in Figure~\ref{fig:medPromptVQAsampleAnswer}. Prompting design templates for a sample brain tumor VQA task and a mammography VQA task are illustrated in Figure~\ref{fig:brainPromptVQAsample} and Figure~\ref{fig:mamPromptVQAsample}, respectively.

\begin{figure}[H]
    \vspace*{-0.3in}
    \centering
    \begin{tcolorbox}[
            colback=promptgray,
            colframe=black!50!black,
            colbacktitle=white!50!black,
            title=\textbf{Prompting Sample from MedXpertQA (Case: MM-1993)},
            enhanced,
            overlay={
                \node[anchor=north east, inner sep=0pt, yshift=-31pt] 
                at (frame.north east) {\includegraphics[width=0.23\textwidth]{images/MM-1993-a.jpeg}};
            } 
            fonttitle=\Large,
            fontupper=\footnotesize,
            center title,
            toptitle=8pt,
            bottomtitle=5pt,
            left=-20pt,
            bottom=-5pt,
        ]
        
    {\linespread{0.85}\selectfont
    \begin{verbatim}
    [
      {
        "role": "system",
        "content": "You are a helpful medical assistant."
      },
      {
        "role": "user",
        "content": [
          {
            "type": "text",
            "text": "
                Q: A 45-year-old man is brought to the emergency department by police after being found 
                unconscious in a store. He is wearing soiled clothing that smells of urine, and his pants
                are soaked in vomit. His medical history includes IV drug use, alcohol use, and fractures
                due to scurvy. He is not on any current medications. Initial vital signs show a temperature
                of 99.5°F (37.5°C), blood pressure of 90/63 mmHg, pulse of 130/min, respirations of 15/min,
                and oxygen saturation of 95% on room air. The patient is treated with IV fluids, thiamine,
                and dextrose, after which he becomes more alert but continues vomiting. Physical examination
                reveals epigastric tenderness, while cardiac and pulmonary exams are unremarkable. A CT scan
                of the abdomen is performed, and laboratory results are as follows:
                 - Serum:                      
                    Na+: 139 mEq/L                    Creatinine: 1.1 mg/dL            
                    Cl-: 102 mEq/L                    Ca2+: 10.2 mg/dL
                    K+: 4.0 mEq/L                     Lipase: 295 U/L
                    HCO3-: 26 mEq/L                   AST: 57 U/L
                    BUN: 20 mg/dL                     ALT: 39 U/L
                    Glucose: 73 mg/dL
                 - Hematology:                      
                    Hemoglobin: 9 g/dL                Hematocrit: 30%
                    Mean corpuscular volume: 120 µm³  Leukocyte count: 8,500/mm³
                    Platelet count: 199,000/mm³                 
                Several hours later, his vital signs improve to a temperature of 99.5°F (37.5°C), blood
                pressure of 110/72 mmHg, pulse of 97/min, respirations of 15/min, and oxygen saturation of
                95% on room air. On examination, suprasternal crepitus is noted, along with blood-stained
                vomitus in the oropharynx. Cardiac and pulmonary findings remain normal, and the lower
                extremities show no abnormalities. What is the most appropriate next step in this
                patient’s management?            
                Answer Choices: 
                    (A) Ondansetron               (B) Folate and vitamin B12 
                    (C) Supportive therapy        (D) Injection of epinephrine 
                    (E) Gastrografin swallow
                A: Let's think step by step."
          },
          {
            "type": "image_url",
            "image_url": {"url": "images/MM-1993-a.jpeg"}
          }
        ]
      },
      {
        "role": "assistant",
        "content": "{ASSISTANT_RATIONALE}"
      },
      {
        "role": "user",
        "content": "Therefore, among A through E, the answer is"
      },
      {
        "role": "assistant",
        "content": "E"
      }
    ]
    \end{verbatim}
    }
    \end{tcolorbox}

    \vspace*{-3mm} 
    \caption{Detailed prompting design template for a representative question taken from the MedXpertQA dataset (Case: MM-1993). The associated image is provided first, followed by the question and labeled answer choices. The model outputs an intermediate reasoning step (\texttt{ASSISTANT\_RATIONALE}) and then a final letter (\texttt{ASSISTANT\_FINAL}) corresponding to its chosen answer. Accuracy was computed solely from \texttt{ASSISTANT\_FINAL}.}
    \label{fig:medPromptVQAsample}
\end{figure}

\begin{figure}[H]
    \centering
    \begin{tcolorbox}[
            colback=promptgray,
            colframe=black!50!black,
            colbacktitle=white!50!black,
            title=\textbf{GPT-5 Reasoning and Final Answer for MedXpertQA (Case: MM-1993)},
            enhanced,
            overlay={
                \node[anchor=north east, inner sep=0pt, yshift=-31pt] 
                at (frame.north east) {\includegraphics[width=0.23\textwidth]{images/MM-1993-a.jpeg}};
            } 
            fonttitle=\Large,
            fontupper=\large,
            center title,
            toptitle=8pt,
            bottomtitle=5pt,
            left=-20pt,
            bottom=-5pt,
        ]
        
    {\linespread{0.85}\selectfont
    \begin{verbatim}
    {
        "role": "assistant",
        "content": "Answer: E) Gastrografin swallow
    
        Reasoning:
        - He likely has acute alcoholic pancreatitis (epigastric 
            pain, elevated lipase, CT changes), which led to re-
            peated vomiting.
        - New suprasternal crepitus with blood-streaked emesis 
            after forceful vomiting strongly suggests esophageal 
            perforation (Boerhaave syndrome), not just a Mallory-
            Weiss tear.
        - The next step to confirm suspected perforation is a water-soluble
            contrast esophagram (Gastrografin swallow). If positive, proceed with
            NPO, broad-spectrum antibiotics, and urgent surgical consultation.
        
        Why not the others:
        - A) Ondansetron: antiemetic alone misses a life-threatening perforation.
        - B) Folate/B12: macrocytosis can be addressed later; not the acute issue.
        - C) Supportive therapy: appropriate for pancreatitis, but new crepitus 
            requires urgent evaluation for perforation.
        - D) Epinephrine: for anaphylaxis or endoscopic hemostasis, not indicated 
            here."
    },
    {
        "role": "assistant",
        "content": "E) Gastrografin swallow"
    }
    \end{verbatim}
    }
    \end{tcolorbox}

    \vspace*{-3mm} 
    \caption{Output of GPT-5 intermediate reasoning step stored in the \texttt{ASSISTANT\_RATIONALE} variable and final answer for MedXpertQA case MM-1993.}
    \label{fig:medPromptVQAsampleAnswer}
\end{figure}

\newpage
\bibliographystyle{elsarticle-num}
\bibliography{refs}

\end{document}